\documentclass[10pt,twocolumn,letterpaper]{article}

\usepackage{cvpr}
\usepackage{times}
\usepackage{epsfig}
\usepackage{graphicx}
\usepackage{amsmath}
\usepackage{amssymb}

\usepackage{algpseudocode}
\usepackage{algorithmicx,algorithm}

\usepackage[pagebackref=true,breaklinks=true,letterpaper=true,colorlinks,bookmarks=false]{hyperref}
\usepackage{makecell}
\usepackage[breaklinks=true,bookmarks=false]{hyperref}

\cvprfinalcopy % *** Uncomment this line for the final submission

\def\cvprPaperID{****} % *** Enter the CVPR Paper ID here

\begin{document}

%%%%%%%%% TITLE
\title{Weakly-Supervised Semantic Segmentation by Iteratively Mining \\Common Object Features}

\author{Xiang Wang$^1$\qquad Shaodi You$^{2,3}$\qquad Xi Li$^1$\qquad Huimin Ma$^{1}$\footnotemark[1]\\
$^1$ Department of Electronic Engineering, Tsinghua University\\
$^2$ DATA61-CSIRO \qquad$^3$ Australian National University\\
{\tt\small \{wangxiang14@mails., lixi16@mails., mhmpub@\}tsinghua.edu.cn, shaodi.you@data61.csiro.au}}

%\and

\maketitle
\pagestyle{empty}  % no page number for the second and the later pages  
\thispagestyle{empty} % no page number for the first page  

%%%%%%%%% ABSTRACT
\begin{abstract}
\hyphenpenalty=8000
\tolerance=2000
Weakly-supervised semantic segmentation under image tags supervision is a challenging task as it directly associates high-level semantic to low-level appearance. 
To bridge this gap, in this paper, we propose an iterative bottom-up and top-down framework which alternatively expands object regions and optimizes segmentation network.
We start from initial localization produced by classification networks. While classification networks are only responsive to small and coarse discriminative object regions, we argue that, these regions contain significant common features about objects. So in the bottom-up step, we mine common object features from the initial localization and expand object regions with the mined features. To supplement non-discriminative regions, saliency maps are then considered under Bayesian framework to refine the object regions. Then in the top-down step, the refined object regions are used as supervision to train the segmentation network and to predict object masks. These object masks provide more accurate localization and contain more regions of object. Further, we take these object masks as initial localization and mine common object features from them. These processes are conducted iteratively to progressively produce fine object masks and optimize segmentation networks.
Experimental results on Pascal VOC 2012 dataset demonstrate that the proposed method outperforms previous state-of-the-art methods by a large margin.

\end{abstract}

\footnotetext[1]{corresponding author}
%%%%%%%%% BODY TEXT

\section{Introduction}\label{sec:introduction}

Weakly-supervised semantic segmentation under image tags supervision is to perform a pixel-wise segmentation of an image, providing only the labels of existing semantic objects in the image.
Because it relies on very slight human labeling, it benefits a number of computer vision tasks, such as object detection~\cite{gidaris2015object} and autonomous driving~\cite{cordts2016cityscapes}.

Weakly-supervised semantic segmentation is, however, very challenging as it directly associates high-level semantic to low-level appearance.
Since only image tags are available, most previous works rely on classification networks to localize objects. However, while no pixel-wise annotation is available, classification networks can only produce inaccurate and coarse discriminative object regions, which can not meet the requirement of pixel-wise semantic segmentation, and thus harms the performance.

\begin{figure}[t] \label{fig:FirstFigure}%\footnotesize
\begin{center}
\includegraphics[width=1\linewidth,trim = 0mm 0mm 0mm 0mm, clip]{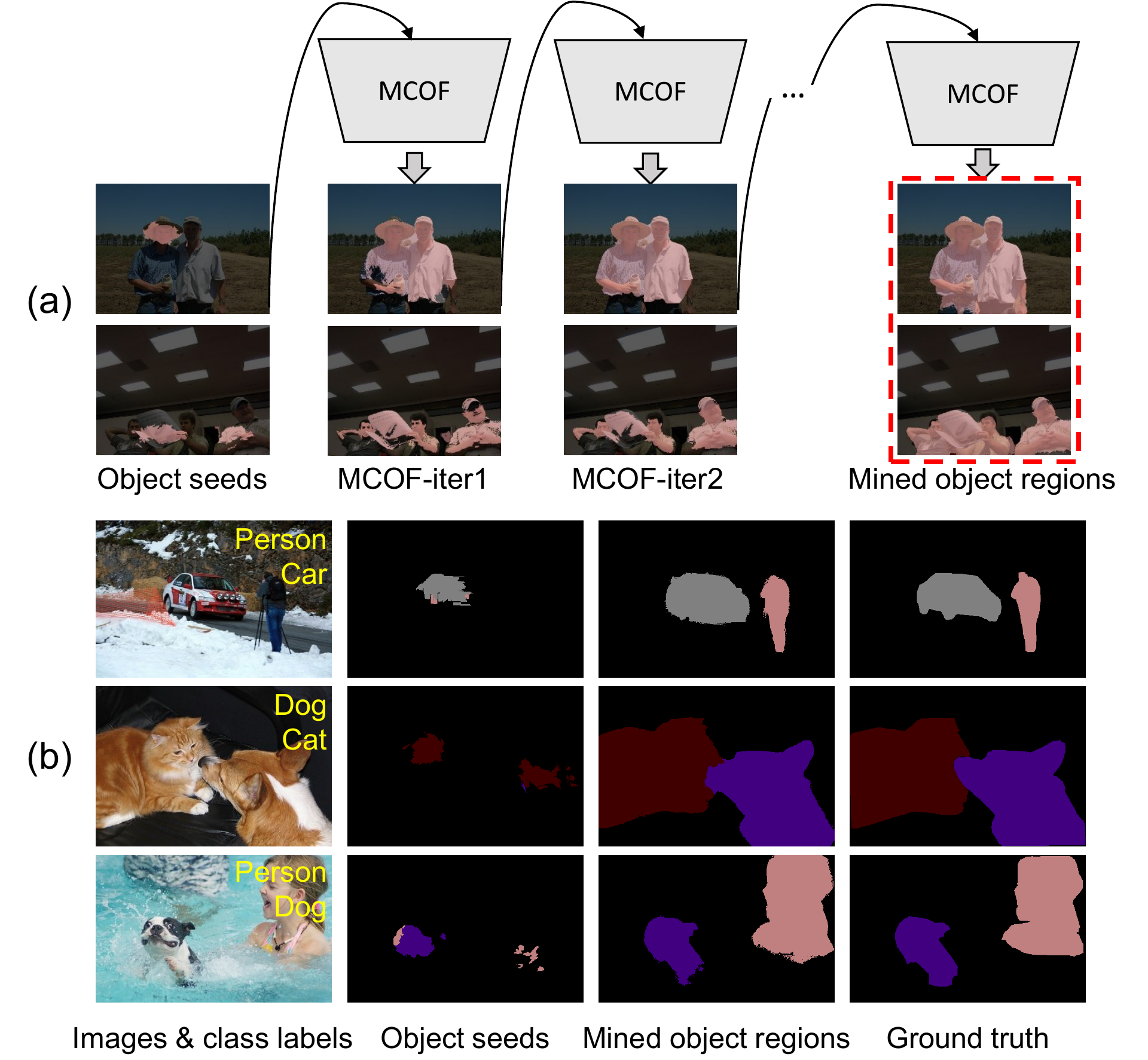}
\vspace{-5mm}
\caption{(a) Illustration of the proposed \textit{MCOF} framework. Our framework iteratively mines common object features and expands object regions. (b) Examples of initial object seeds and our mined object regions. Our method can tolerate inaccurate initial localization and produce quite satisfactory results.}
\end{center}
\vspace{-8mm}
\end{figure}

To address this issue, in this paper, we propose an iterative bottom-up and top-down framework, which tolerates inaccurate initial localization by Mining Common Object Features (\textit{MCOF}) from initial localization to progressively expand object regions. Our motivation is, though the initial localization produced by classification network is coarse, it gives certain discriminative regions of objects, these regions contain important knowledge about objects, \ie common object features. For example, as shown in Figure~\ref{fig:FirstFigure} (a), some images may locate hands of person, while other images may locate heads. Given a set of training images, we can learn common object features from them to predict regions of whole object.
So in the bottom-up step, we take the initial object localization as object seeds and mine common object features from them to expand object regions. Then in the top-down step, we train segmentation network using the mined object regions as supervision to predict fine object masks. 
The predicted object masks contain more regions of objects, which are more accurate and provide more training samples of objects, so we can further mine common object features from them.
And the processes above are conducted iteratively to progressively produce fine object regions and optimize segmentation networks.
With iterations, inaccurate regions in the initial localization are progressively corrected, so our method is robust and can tolerate inaccurate initial localization. Figure~\ref{fig:FirstFigure} (b) shows some examples in which the initial localization is very coarse and inaccurate, while our method can still produce satisfactory results.

Concretely, we first train an image classification network and localize discriminative regions of object using Classification Activation Maps (CAM)~\cite{zhou2016learning}. Images are then segmented into superpixel regions and are assigned with class labels using CAM, these regions are called initial object seeds. The initial object seeds contain certain key parts of objects, 
so in bottom-up step, we mine common object features from them and then expand object regions. We achieve this by training a region classification network and use the trained network to predict object regions. While these regions may still only focus on key part regions of objects, to supplement non-discriminative regions, saliency-guided refinement method is proposed which considers both the expanded object regions and saliency maps under Bayesian framework. 
Then in top-down step, we train segmentation network using the refined object regions as supervision to predict segmentation masks.
With the aforementioned  procedure, we can get segmentation masks which contain more complete object regions and are much more accurate than the initial object seeds. We further take the segmentation masks as object seeds, and conduct the processes iteratively. 
With iterations, the proposed \textit{MCOF} framework progressively produces more accurate object regions and enhances the performance of the segmentation network.
The final trained segmentation network is applied for inference.

The main contributions of our work are three-fold:
\vspace{-2mm}
\begin{itemize}
\item We propose an iterative bottom-up and top-down framework which tolerates inaccurate initial localization by iteratively mining common object features to progressively produce accurate object masks and optimize segmentation network.\vspace{-2mm}
\item Saliency-guided refinement method is proposed to supplement non-discriminative regions which are ignored in initial localization.\vspace{-2mm}
\item Experiments on PASCAL VOC 2012 segmentation dataset demonstrate that our method outperforms previous methods and achieves state-of-the-art performance.\vspace{-2mm}
\end{itemize}

%In this paper, we aims to consider properties of objects by mining common features. As illustrated in Figure~\ref{fig:pipeline}, we initialize our system by generating object seeds from localization map produced in classification networks~\cite{zhou2016learning}. This provides our system to roughly locate the objects on image. Later, based on the object seeds, our system gradually learns the appearance of objects by mining common appearance features. Through the iterative mining, the seeds gradually expanded to the semantic areas to supervise segmentation networks.

\begin{figure*}[t] %\footnotesize
\begin{center}
\includegraphics[width=0.95\linewidth,trim = 0mm 0mm 0mm 0mm, clip]{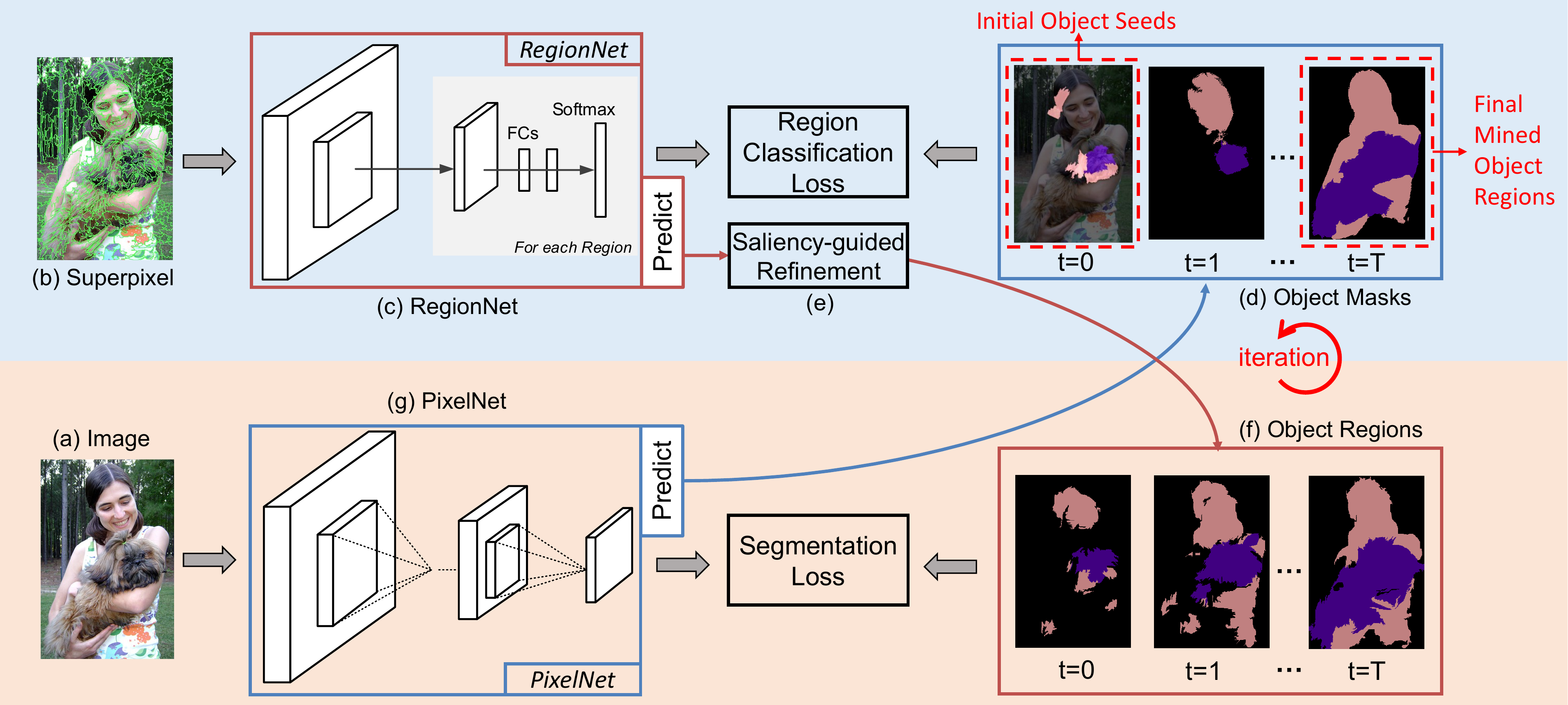}
\vspace{1mm}
\caption{Pipeline of the proposed \textit{MCOF} framework. At first (t=0), we mine common object features from initial object seeds. We segment (a) image into (b) superpixel regions and train the (c) region classification network \textit{RegionNet} with the (d) initial object seeds. We then re-predict the training images regions with the trained \textit{RegionNet} to get object regions. While the object regions may still only focus on discriminative regions of object, we address this by (e) saliency-guided refinement to get (f) refined object regions. The refined object regions are then used to train the (g) \textit{PixelNet}. With the trained \textit{PixelNet}, we re-predict the (d) segmentation masks of training images, are then used them as supervision to train the \textit{RegionNet}, and the processes above are conducted iteratively. With the iterations, we can mine finer object regions and the \textit{PixelNet} trained in the last iteration is used for inference.}
\label{fig:pipeline}
\end{center}
\vspace{-8mm}
\end{figure*}

\section{Related Work}\label{sec:relatedwork}
In this section, we introduce both fully-supervised and weakly-supervised semantic segmentation networks which are related to our work.

\subsection{Fully-Supervised Semantic Segmentation}
Fully-supervised methods acquire a large number of pixel-wise annotations, according to the process mode, they can be categorized as region-based and pixel-based networks.

Region-based networks take images as a set of regions and extract features of them to predict their labels. Mostajabi \etal~\cite{mostajabi2015feedforward} proposed zoom-out features which combines features of local, proximal, distant neighboring superpixels and the entire scene to classify each superpixel.

Pixel-based networks take the entire image as input and predict pixel-wise labels end-to-end with fully convolutional layers. Long~\etal~\cite{long2015fully} proposed fully convolutional network (FCN) and skip architecture to produce accurate and detailed semantic segmentation. Chen~\etal~\cite{chen2014semantic} proposed DeepLab which introduces ``hole algorithm'' to enlarge the receptive field with lower stride to produce denser segmentation. A large number of works~\cite{chen2016semantic, noh2015learning, xia2015zoom} have been proposed based on FCN and DeepLab.

Pixel-based networks have been proved to be more powerful than Region-based networks for semantic segmentation. However, in this paper, we take advantages of both kinds of networks. We show that region-based networks are powerful in learning common features of objects and thus can produce fine object regions as supervision to train pixel-based networks.

\subsection{Weakly-Supervised Semantic Segmentation}
While fully-supervised methods require a large number of pixel-wise annotation which is very expensive, recent advances have exploited semantic segmentation with weak supervision, including bounding box~\cite{dai2015boxsup, papandreou2015weakly, khoreva2017simple}, scribble~\cite{lin2016scribblesup} and image-level labels~\cite{pathakICLR15, pinheiro2015image, saleh2016built, papandreou2015weakly, wei2016stc, kolesnikov2016seed, qi2016augmented, wei2017object}. In this paper, we only focus on the weakest supervision, \textit{i.e.}, image-level supervision.

In image-level weakly-supervised semantic segmentation, since only image tags are available, most methods are based on classification methods, and these methods can be coarsely classified into two categories: \textit{MIL-based} methods, which directly predict segmentation masks with classification networks; and \textit{localization-based} methods, which utilize classification networks to produce initial localization and use them to supervise segmentation networks. 

Multi-instance learning (MIL) based methods~\cite{pathakICLR15, pinheiro2015image, kolesnikov2016seed, saleh2016built, durand2017wildcat} formulate weakly-supervised learning as a MIL framework in which each image is known to have at least one pixel belonging to a certain class, and the task is to find these pixels. Pinheiro~\etal~\cite{pinheiro2015image} proposed Log-Sum-Exp (LSE) to pool the output feature maps into image-level labels, so that the network can be trained end-to-end as a classification task. Kolesnikov~\etal~\cite{kolesnikov2016seed} proposed global weighted rank pooling (GWRP) method which gives more weights to promising location in the last pooling layer. 
However, while MIL-based methods can locate discriminative object regions, they suffer from coarse object boundaries and thus the performance is not satisfactory.

Localization-based methods~\cite{papandreou2015weakly, wei2016stc, kolesnikov2016seed, qi2016augmented, wei2017object} aim to generate initial object localization from weak labels and then use it as supervision to train segmentation networks. 
%Wei \textit{et.al}~\cite{wei2016stc} proposed a simple to complex framework, they first trained network from simple images with saliency maps, then they iteratively used the predict results of previous network as supervision to train the next enhanced network. 
Kolesnikov~\etal~\cite{kolesnikov2016seed} used localization cues generated from classification networks as a kind of supervision, they also proposed classification loss and boundary-aware loss to consider class and boundary constrain. Wei~\etal~\cite{wei2017object} proposed adversarial erasing method to progressively mine object region with classification network.
While Wei~\etal~\cite{wei2017object} also aims to expand object regions from the initial localization. They rely on the classification network to sequentially produce the most discriminative regions in erased images. It will cause error accumulation and the mined object regions will have coarse object boundary.
%Among them, \cite{kolesnikov2016seed} and ~\cite{wei2017object} localize object seeds from classification network similar to ours. However, Kolesnikov~\etal~\cite{kolesnikov2016seed} directly based on the coarse object map produced by classification network~\cite{zhou2016learning}, thus the performance is limited. And the adversarial erasing method in~\cite{wei2017object} will cause error accumulation and make the object region inaccurate. 
The proposed \textit{MCOF} method mines common object features from coarse object seeds to predict finer segmentation masks, and then iteratively mines features from the predicted masks. Our method progressively expands object regions and corrects inaccurate regions, which is robust to noise and thus can tolerate inaccurate initial localization. By taking advantages of superpixel, the mined object regions will have clear boundary.

\begin{algorithm}[t]
    \hyphenation{object}
    \caption{Framework of the proposed \textit{MCOF}} 
    \hspace*{0.02in} {\bf Input:} 
    Training images $\mathcal{I} $ and Superpixel regions $\mathcal{R}$ \\
    \hspace*{0.02in} {\bf Initialize:} Generate initial object seeds $\mathcal{S}$, $t$ = 0. 
    \begin{algorithmic}[1]
        \While{iteration is effective}
        \State Train the \textit{RegionNet} with $\mathcal{R}$ and $\mathcal{S}$
        \State Predict with the trained \textit{RegionNet} to get object \mbox{\quad{ }{ }}regions $\mathcal{O}$.
        \If{$t$ == 0}
        \State Refine object regions $\mathcal{O}$ with saliency maps to \mbox{\quad\quad{ }{ } } get refined object regions $\mathcal{O}^R$
        \Else
        \State $\mathcal{O}^R \gets \mathcal{O}$
        \EndIf
        \State Train the \textit{PixelNet} with $\mathcal{I}$ and $\mathcal{O}^R$
        \State Predict with the trained \textit{PixelNet} to get object \mbox{\quad{ }{ }}masks $\mathcal{M}$
        \State Update $\mathcal{S} \gets \mathcal{M}$, $t \gets t + 1$.
        \EndWhile
    \end{algorithmic}
    \hspace*{0.02in} {\bf Output:} Mined object masks $\mathcal{M}$ and the trained \textit{PixelNet}
    \label{algo}
\end{algorithm}

\section{Architecture of the Proposed MCOF}
Classification networks can only produce coarse and inaccurate discriminative object localization, which are far from the requirement of pixel-wise semantic segmentation. To address this issue, in this paper, we argue that, though the initial object localization is coarse, it contains important features about objects. So we propose to mine common object features from initial object seeds to progressively correct inaccurate regions and produce fine object regions to supervise segmentation network. 

As shown in Figure~\ref{fig:pipeline}, our framework consists of two iterative steps: bottom-up step and top-down step.
The bottom-up step mines common object features from object seeds to produce fine object regions, and the top-down step uses the produced object regions to train weakly-supervised segmentation network. The predicted segmentation masks contain more complete object regions than initial. We then take them as object seeds to mine common object features and the processes are conducted iteratively to progressively correct inaccurate regions and produce fine object regions. 

Note that, in the first iteration, the initial object seeds only contain discriminative regions, after mining common object features, some non-discriminative regions are still missing. To address this, we propose to incorporate saliency maps with the mined object regions.
After the first iteration, the segmented masks contain much more object regions and are more accurate, while the accuracy of saliency maps are also limited, so in the later iterations, the saliency maps are not used to prevent introducing additional noise.
The overall procedure is summarized as Algorithm~\ref{algo}.

It is worth noting that the iterative processes are only applied in the training stage, for inference, only the segmentation network of the last iteration is utilized, so the inference is efficient.

\section{Mining Common Object Features}\label{sec:method}
\subsection{Initial Object Seeds}~\label{sec:seeds}
\begin{figure}[t] %\footnotesize
\begin{center}
\includegraphics[width=0.96\linewidth,trim = 0mm 0mm 0mm 0mm, clip]{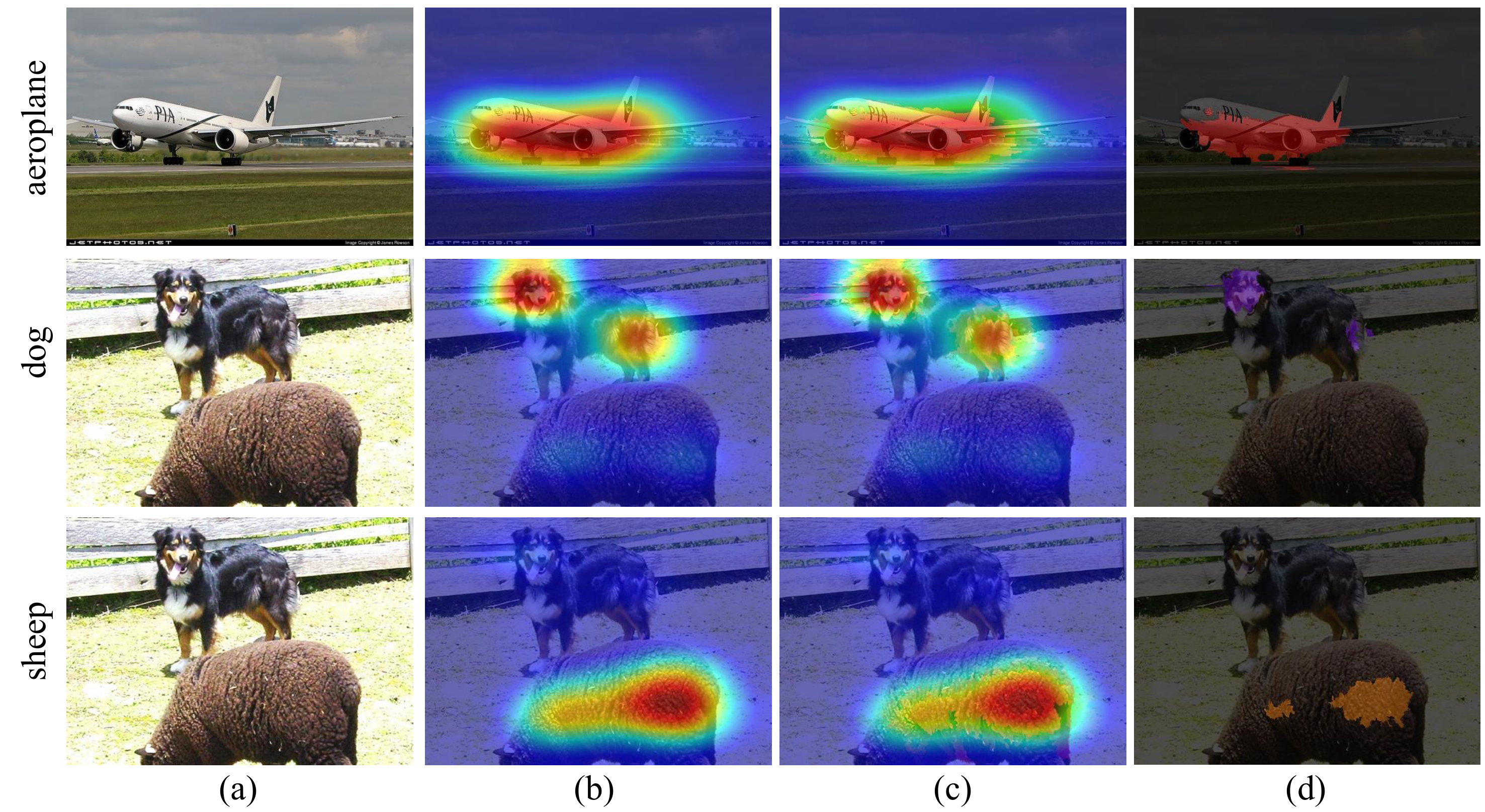}
\vspace{-0mm}
\caption{Examples of generating initial object seeds from object heatmaps. (a) Images, (b) object heatmaps of CAM~\cite{zhou2016learning}, (c) object heatmaps averaged in each superpixel, (d) initial object seeds.}
\label{fig:cam_example}
\end{center}
\vspace{-8mm}
\end{figure}

To get initial object localization, we train a classification network and use CAM method~\cite{zhou2016learning} to produce heatmap of each object. As shown in Figure~\ref{fig:cam_example}, the heatmap is very coarse, to localize discriminative regions of objects, first, we segment images into superpixel regions using graph-based segmentation method~\cite{felzenszwalb2004efficient} and average the heatmap within each region. We observe that the CAM map usually has several center regions with low-confidence regions surrounding them, and the center regions are mostly the key part of objects. So for each heatmap, we select its local maximum region as initial seeds. However, this may miss lots of regions, so regions with heatmap larger than a threshold are also selected as initial seeds. Some examples are shown in Figure~\ref{fig:cam_example}.

\subsection{Mining Common Object Features from Initial Object Seeds}~\label{sec:regionnet}
\begin{figure*}[t] %\footnotesize
\begin{center}
\includegraphics[width=1\linewidth,trim = 0mm 0mm 0mm 0mm, clip]{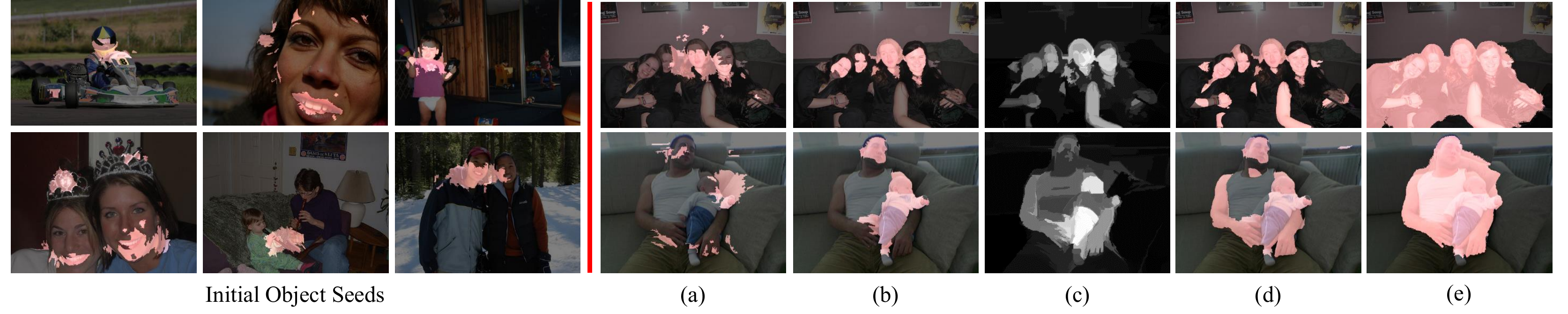}
\vspace{-6mm}
\caption{Left: examples of object seeds. They give us features of objects of different locations. However, they mainly focus on key parts which are helpful for recognition. Right: (a) initial object seeds, (b) object masks predicted by \textit{RegionNet}, (c) saliency map, (d) refined object regions via Bayesian framework, (e) segmentation results of \textit{PixelNet}.}
\label{fig:seeds}
\end{center}
\vspace{-8mm}
\end{figure*}

The initial object seeds are too coarse to meet the requirement of semantic segmentation, however, they contain discriminative regions of objects. For example, as shown in Figure~\ref{fig:seeds}, one image may locate hands of a person, while another may give the location of face. We argue that, regions of same class have some shared attributions, namely, common object features. So given a set of training images with seed regions, we can learn common object features from them and predict the whole regions of object, thus to expand object regions and suppress noisy regions. We achieve this by training a region classification network, named \textit{RegionNet}, using the object seeds as training data.

Formally, given $N$ training images~$\mathcal{I} = \{I_i\}_{i=1}^N$, we first segment them into superpixel regions~$\mathcal{R} = \{R_{i,j}\}_{i=1,j=1}^{N,n_i}$ using graph-based segmentation method~\cite{felzenszwalb2004efficient}, where $n_i$ is the number of superpixel regions of the image $I_i$. In Sec~\ref{sec:seeds}, we have got initial object seeds, with them, we can give labels for superpixel regions~$\mathcal{R}$ and denote them as $\mathcal{S} = \{S_{i,j}\}_{i=1,j=1}^{N,n_i}$, where $S_{i,j}$ is one-hot encoding with $S_{i,j}(c) = 1$ and others as 0 if $R_{i,j}$ belongs to class $c$. Based on training data $\mathcal{D}=\{(R_{i,j}, S_{i,j})\}_{i=1,j=1}^{N,n_i}$, our goal is to train a region classification network~$f^r(R;\theta_{r})$ parameterized by $\theta_{r}$ to model the probability of region $R_{i,j}$ being class label $c$ , namely, $f^r_c(R_{i,j}|\theta_{r}) = p(y=c|R_{i,j})$.

We achieve this with the efficient mask-based Fast R-CNN framework~\cite{girshick2015fast, wang2016edge, wang2018edge}. In this framework, we take \textit{external rectangle} of each region as the \textit{RoI} of the original Fast R-CNN framework. In the \textit{RoI} pooling layer, features inside superpixel regions are pooled while features inside the external rectangle but outside the superpixel regions are pooled as zero. To train this network, we minimize the cross-entropy loss function:
\begin{equation}
\mathcal{L}_r = -\sum\limits_{i,j,c}S_{i,j}(c)log(f^r_c(R_{i,j}|\theta_{r})).
\end{equation}

By training the \textit{RegionNet}, common object features can be mined from the initial object seeds. We then use the trained network to predict the label of each region of the training images.
In the prediction, some incorrect regions and regions initially labeled as background can be classified correctly, thus to expand object regions. Some examples are shown in Figure~\ref{fig:seeds} (a) and (b), we can see that object regions predicted by \textit{RegionNet} contain more regions of objects and some noisy regions in initial object seeds are corrected. In this paper, we call these regions as object regions and denote them as $\mathcal{O}=\{O_i\}_{i=1}^N$.

Note that since we have the class labels of training images, we can remove wrong predictions and label them as background. This will guarantee that the produced object regions do not contain any non-existent class, which is important for training the following segmentation network.

\begin{figure}[t] %\footnotesize
\begin{center}
\includegraphics[width=1\linewidth,trim = 0mm 0mm 0mm 0mm, clip]{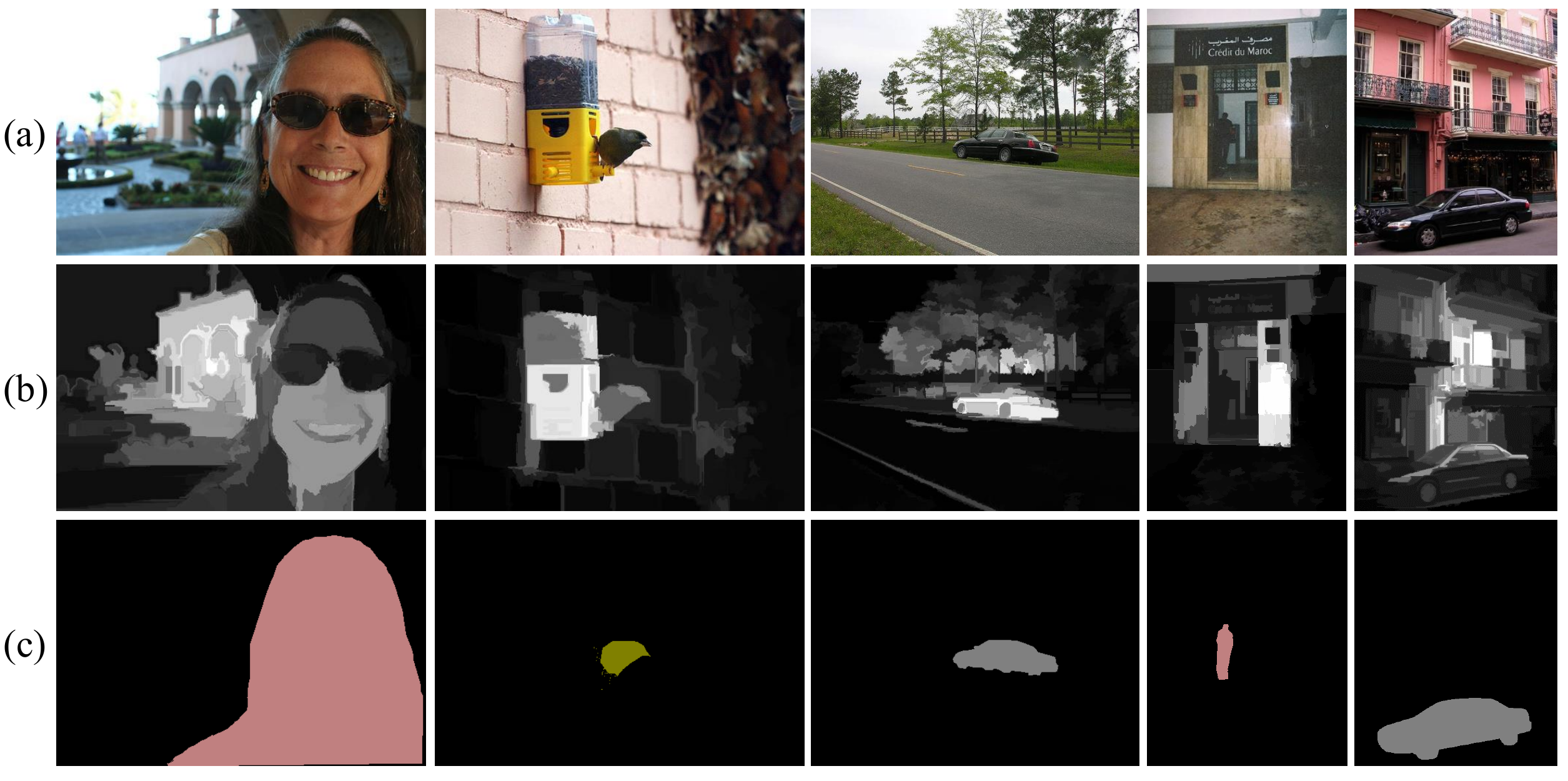}
\vspace{-4mm}
\caption{For images with single object class, salient object regions may not be consistent with semantic segmentation. In addition, they may be inaccurate and may locate other objects which are not included in semantic segmentation datasets. (a) Images, (b) saliency map of DRFI~\cite{jiang2013salient}, (c) semantic segmentation.}
\label{fig:bad_saliency}
\end{center}
\vspace{-6mm}
\end{figure}

\subsection{Saliency-Guided Object Region Supplement}~\label{sec:saliency}
Note that the \textit{RegionNet} is learned from the initial seed regions which mainly contain key regions of objects. With the \textit{RegionNet}, the object regions can be expanded while there still exists some regions that are ignored. For example, the initial seed regions mainly focus on heads and hands of a person, while other regions, such as the body, are often ignored. After expanding by \textit{RegionNet}, some regions of the body are still missing (Figure~\ref{fig:seeds} (b)).

To address this issue, we propose to supplement object regions by incorporating saliency maps for images with single object class. Note that we do not directly use saliency map as initial localization as previous works~\cite{wei2016stc}, since in some cases, salient object may not be the object class we need in semantic segmentation, and the saliency map itself also contains noisy regions which will affect the localization accuracy. Some examples are shown in Figure~\ref{fig:bad_saliency}.

We address this by proposing saliency-guided object region supplement method which considers both the mined object regions and saliency maps under Bayesian framework. In Sec~\ref{sec:regionnet}, we have mined object regions which contains key parts of objects. Based on these key parts, we aim to supplement object regions with saliency maps. Our idea is, for a region with high saliency value, if it's similar with the mined object objects, then it is more likely to be part of that object. We can formulate the above hypothesis under Bayesian optimization~\cite{xie2013bayesian, wang2016geodesic} as:
\begin{equation}
p(obj|\mbox{\boldmath{$v$}}) = \frac{p(obj)p(\mbox{\boldmath{$v$}}|obj)}{p(obj)p(\mbox{\boldmath{$v$}}|obj) + p(bg)p(\mbox{\boldmath{$v$}}|bg)},
\end{equation}
where $p(obj)$ is the saliency map, and $p(bg) = 1 - p(obj)$, $p(\mbox{\boldmath{$v$}}|obj)$ and $p(\mbox{\boldmath{$v$}}|bg)$ are the feature distribution at object regions and background regions, $\mbox{\boldmath{$v$}}$ is the feature vector, $p(obj|\mbox{\boldmath{$v$}})$ is the refined object map which represents the probability of region with feature $\mbox{\boldmath{$v$}}$ being object. By binarizing the refined object map $p(obj|\mbox{\boldmath{$v$}})$ with a CRF~\cite{koltun2011efficient}, we can get refined object regions which incorporate saliency maps to supplement the original object regions. In our work, we use saliency map of the DRFI method~\cite{jiang2013salient} as in~\cite{wei2016stc}. 

Some examples are shown in Figure~\ref{fig:seeds}, by incorporating saliency maps, more object regions are included.
In this paper, we call these regions as refined object regions and denote them as $\mathcal{O}^R=\{O_i^R\}_{i=1}^N$.

\begin{figure*}[!ht] %\footnotesize
\begin{center}
\includegraphics[width=1\linewidth,trim = 0mm 0mm 0mm 0mm, clip]{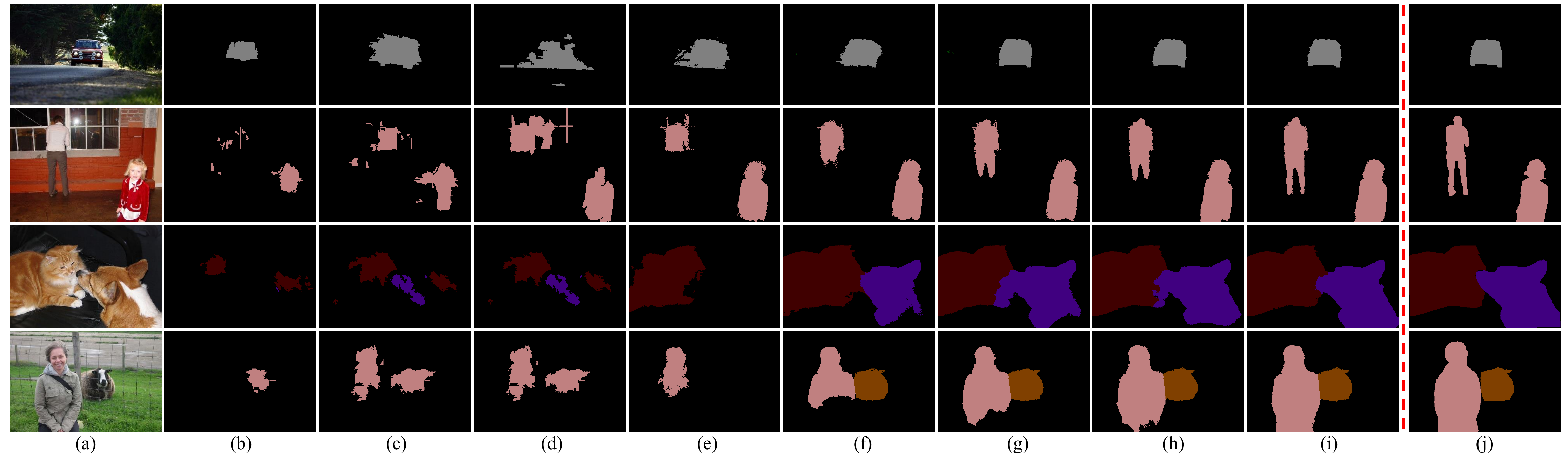}
\vspace{-5mm}
\caption{Intermediate results of the proposed framework. (a) Image, (b) initial object seeds, (c) expanded object regions predicted by \textit{RegionNet}, (d) saliency-guided refined object regions. Note that, the saliency-guided refinement is only applied to images with single class, for images with multiple classes (3rd and 4th rows), the object regions remain unchanged. Segmentation results of \textit{PixelNet} in (e) 1st, (f) 2nd, (g) 3rd, (h) 4th and (i) 5th iteration, (j) ground truth.}
\label{fig:iterationsresults}
\end{center}
\vspace{-8mm}
\end{figure*}

\section{Iterative Learning Framework}~\label{sec:segment}
%\subsection{Weakly-Supervised Semantic Segmentation}~\label{sec:segment}
The refined object regions give us some reliable localization of object, we can use them as supervision to train the weakly-supervised semantic segmentation network. 
While previous works~\cite{kolesnikov2016seed, wei2017object, durand2017wildcat} rely on both localization cues and class labels to design and train segmentation network, in our work, we have removed wrong class regions in the previous \textit{RegionNet}, thus the refined object regions do not contain any wrong class.
%Different from previous methods~\cite{kolesnikov2016seed, wei2017object, durand2017wildcat} which utilized both localization and label cues in segmentation network, in our work, the refine object masks do not contain any wrong class as we have removed it in the previous region classification network. 
So we can only use the localization cues as supervision, this is completely compatible with fully-supervised framework, and thus we can benefit from existing fully-supervised architecture. In this paper, we utilize the popular DeepLab-LargeFOV model~\cite{chen2014semantic} as the basic network of our segmentation network, named \textit{PixelNet}.

Formally, given the training images~$\mathcal{I} = \{I_i\}_{i=1}^N$ and corresponding refined object regions~$\mathcal{O}^R=\{O_i^R\}_{i=1}^N$, our goal is to train the segmentation network~$f^s(I;\theta_{s})$ parameterized by $\theta_{s}$ to model the probability that location $u$ being the class label $c$, namely, $f^s_{u,c}(I|\theta_s)=p(y_u=c|I)$. The loss function is the cross-entropy loss which encourages the predictions to match our refined object regions:
\begin{equation}
\mathcal{L}_s = -\frac{1}{\sum_{c=1}^{C}|S_c|}\sum\limits_{c=1}^C\sum\limits_{u\in{S_c}} log(f^s_{u,c}(I|\theta_{s})),
\end{equation}
where $C$ is the number of classes and $S_c$ is a set of locations that are labelled with class $c$ in the supervision.

The supervision cues, namely, the object regions, is produced by the region classification network, it only considers features inside each region. While in the \textit{PixelNet}, the whole image is considered and thus the context information is utilized. Using the trained \textit{PixelNet} to predict the segmentation masks of the training images, the segmentation masks will further include more object regions. Some examples are shown in Figure~\ref{fig:seeds}, we can see that the predicted segmentation masks locate more regions of objects and suppress the noisy regions in the previous steps.

Further, we take the predicted segmentation masks as object seeds and conduct the processes above iteratively.
With iterations, more robust common object features can be mined thus to produce finer object regions, and the segmentation network is progressively optimized with better supervision.
Figure~\ref{fig:iterationsresults} shows the results with iterations. With iterations, the object regions are expanded and the inaccurate regions are corrected, so the segmentation results become more and more accurate. Finally, we use the trained \textit{PixelNet} of the last iteration for inference and evaluate it in the experiment section.

\section{Experiments}\label{sec:experiment}
\subsection{Setup}

We evaluate the proposed \textit{MCOF} framework on the PASCAL VOC $2012$ image segmentation benchmark~\cite{everingham2010pascal}~\footnote{The models and results are available at~https://wangxiang10.github.io/}. The dataset contains $20$ object classes and $1$ background class. For the segmentation task, it contains $1464$ training, $1449$ validation and $1456$ test images. Following previous works~\cite{kolesnikov2016seed, qi2016augmented, wei2017object}, we use the augmentation data~\cite{hariharan2011semantic} which contains 10,582 images as training set.
We evaluate our method and compare with other methods on validation and test sets for segmentation task in terms of intersection-over-union averaged on all $21$ classes (mIoU).

\begin{table*}\footnotesize \setlength{\tabcolsep}{2.1pt}
\begin{tabular}{p{3.2cm}ccccccccccccccccccccccc}
\Xhline{1.0pt}
    Method &       \rotatebox{90}{bkg} & \rotatebox{90}{plane} &        \rotatebox{90}{bike} &       \rotatebox{90}{bird} &       \rotatebox{90}{boat} &       \rotatebox{90}{bottle} &       \rotatebox{90}{bus} &        \rotatebox{90}{car} &        \rotatebox{90}{cat} &      \rotatebox{90}{chair} &        \rotatebox{90}{cow} &      \rotatebox{90}{table} &        \rotatebox{90}{dog} &      \rotatebox{90}{horse} &       \rotatebox{90}{motor} &       \rotatebox{90}{person} &      \rotatebox{90}{plant} &      \rotatebox{90}{sheep} &       \rotatebox{90}{sofa} &      \rotatebox{90}{train} &         \rotatebox{90}{tv} & {\bf mIoU} \\
\hline

CCNN (ICCV'15)~\cite{pathak2015constrained} &      68.5  &      25.5  &      18.0  &      25.4  &      20.2  &      36.3  &      46.8  &      47.1  &      48.0  &      15.8  &      37.9  &      21.0  &      44.5  &      34.5  &      46.2  &      40.7  &      30.4  &      36.3  &      22.2  &      38.8  &      36.9  &      35.3  \\

EM-Adapt (ICCV'15)~\cite{papandreou2015weakly} &          - &          - &          - &          - &          - &          - &          - &          - &          - &          - &          - &          - &          - &          - &          - &          - &          - &          - &          - &          - &          - &      38.2  \\

MIL-sppxl (CVPR'15)~\cite{pinheiro2015image} &      77.2  &      37.3  &      18.4  &      25.4  &      28.2  &      31.9  &      41.6  &      48.1  &      50.7  &      12.7  &      45.7  &      14.6  &      50.9  &      44.1  &      39.2  &      37.9  &      28.3  &      44.0  &      19.6  &      37.6  &      35.0  &      36.6  \\

STC (PAMI'16)~\cite{wei2016stc} &      84.5  &      68.0  &      19.5  &      60.5  &      42.5  &      44.8  &      68.4  &      64.0  &      64.8  &      14.5  &      52.0  &      22.8  &      58.0  &      55.3  &      57.8  &      60.5  &      40.6  &      56.7  &      23.0  &      57.1  &      31.2  &      49.8  \\

DCSM (ECCV'16)~\cite{shimoda2016distinct} &      76.7  &      45.1  &      24.6  &      40.8  &      23.0  &      34.8  &      61.0  &      51.9  &      52.4  &      15.5  &      45.9  &      32.7  &      54.9  &      48.6  &      57.4  &      51.8  &      38.2  &      55.4  &      32.2  &      42.6  &      39.6  &      44.1  \\

BFBP (ECCV'16)~\cite{saleh2016built} &      79.2  &      60.1  &      20.4  &      50.7  &      41.2  &      46.3  &      62.6  &      49.2  &      62.3  &      13.3  &      49.7  &      38.1  &      58.4  &      49.0  &      57.0  &      48.2  &      27.8  &      55.1  &      29.6  &      54.6  &      26.6  &      46.6  \\

AF-SS (ECCV'16)~\cite{qi2016augmented} &          - &          - &          - &          - &          - &          - &          - &          - &          - &          - &          - &          - &          - &          - &          - &          - &          - &          - &          - &          - &          - &      52.6  \\

SEC (ECCV'16)~\cite{kolesnikov2016seed} &      82.2  &      61.7  &      26.0  &      60.4  &      25.6  &      45.6  &      70.9  &      63.2  &      72.2  &      20.9  &      52.9  &      30.6  &      62.8  &      56.8  &      63.5  &      57.1  &      32.2  &      60.6  &      32.3  &      44.8  &      42.3  &      50.7  \\

CBTS (CVPR'17)~\cite{roy2017combining} &      85.8  &      65.2  &      29.4  &      63.8  &      31.2  &      37.2  &      69.6  &      64.3  &      76.2  &      21.4  &      56.3  &      29.8  &      68.2  &      60.6  &      66.2  &      55.8  &      30.8  &      66.1  &      34.9  &      48.8  &      47.1  &      52.8  \\

AE-PSL (CVPR'17)~\cite{wei2017object} &          - &          - &          - &          - &          - &          - &          - &          - &          - &          - &          - &          - &          - &          - &          - &          - &          - &          - &          - &          - &          - &      55.0  \\

\hline
\textit{Ours}:\\

MCOF-VGG16 &      85.8  &      74.1  &      23.6  &      66.4  &      36.6  &      62.0  &      75.5  &      68.5  &      78.2  &      18.8  &      64.6  &      29.6  &      72.5  &      61.6  &      63.1  &      55.5  &      37.7  &      65.8  &      32.4  &      68.4  &      39.9  &       \textbf{56.2} \\

MCOF-ResNet101 &      87.0  &      78.4  &      29.4  &      68.0  &      44.0  &      67.3  &      80.3  &      74.1  &      82.2  &      21.1  &      70.7  &      28.2  &      73.2  &      71.5  &      67.2  &      53.0  &      47.7  &      74.5  &      32.4  &      71.0  &      45.8  &       \textbf{60.3} \\

\Xhline{1.0pt}
\end{tabular} 
\vspace{0mm} 
\caption{Comparison of weakly supervised semantic segmentation methods on PASCAL VOC 2012 \textit{val} set.}
\label{tab:val}
\end{table*}

% Table generated by Excel2LaTeX from sheet 'Sheet6'
\begin{table*}\footnotesize \setlength{\tabcolsep}{2.1pt}
\begin{tabular}{p{3.2cm}ccccccccccccccccccccccc}
\Xhline{1.0pt}
    Method &       \rotatebox{90}{bkg} & \rotatebox{90}{plane} &        \rotatebox{90}{bike} &       \rotatebox{90}{bird} &       \rotatebox{90}{boat} &       \rotatebox{90}{bottle} &       \rotatebox{90}{bus} &        \rotatebox{90}{car} &        \rotatebox{90}{cat} &      \rotatebox{90}{chair} &        \rotatebox{90}{cow} &      \rotatebox{90}{table} &        \rotatebox{90}{dog} &      \rotatebox{90}{horse} &       \rotatebox{90}{motor} &       \rotatebox{90}{person} &      \rotatebox{90}{plant} &      \rotatebox{90}{sheep} &       \rotatebox{90}{sofa} &      \rotatebox{90}{train} &         \rotatebox{90}{tv} & {\bf mIoU} \\
\hline
CCNN (ICCV'15)~\cite{pathak2015constrained} &      70.1  &      24.2  &      19.9  &      26.3  &      18.6  &      38.1  &      51.7  &      42.9  &      48.2  &      15.6  &      37.2  &      18.3  &      43.0  &      38.2  &      52.2  &      40.0  &      33.8  &      36.0  &      21.6  &      33.4  &      38.3  &      35.6  \\

EM-Adapt (ICCV'15)~\cite{papandreou2015weakly} &      76.3  &      37.1  &      21.9  &      41.6  &      26.1  &      38.5  &      50.8  &      44.9  &      48.9  &      16.7  &      40.8  &      29.4  &      47.1  &      45.8  &      54.8  &      28.2  &      30.0  &      44.0  &      29.2  &      34.3  &      46.0  &      39.6  \\

MIL-sppxl (CVPR'15)~\cite{pinheiro2015image} &      74.7  &      38.8  &      19.8  &      27.5  &      21.7  &      32.8  &      40.0  &      50.1  &      47.1  &       7.2  &      44.8  &      15.8  &      49.4  &      47.3  &      36.6  &      36.4  &      24.3  &      44.5  &      21.0  &      31.5  &      41.3  &      35.8  \\

STC (PAMI'16)~\cite{wei2016stc} &      85.2  &      62.7  &      21.1  &      58.0  &      31.4  &      55.0  &      68.8  &      63.9  &      63.7  &      14.2  &      57.6  &      28.3  &      63.0  &      59.8  &      67.6  &      61.7  &      42.9  &      61.0  &      23.2  &      52.4  &      33.1  &      51.2  \\

DCSM (ECCV'16)~\cite{shimoda2016distinct}  &      78.1  &      43.8  &      26.3  &      49.8  &      19.5  &      40.3  &      61.6  &      53.9  &      52.7  &      13.7  &      47.3  &      34.8  &      50.3  &      48.9  &      69.0  &      49.7  &      38.4  &      57.1  &      34.0  &      38.0  &      40.0  &      45.1  \\

%&          - &          - &          - &          - &          - &          - &          - &          - &          - &          - &          - &          - &          - &          - &          - &          - &          - &          - &          - &          - &          - &      52.7  \\

BFBP (ECCV'16)~\cite{saleh2016built} &      80.3  &      57.5  &      24.1  &      66.9  &      31.7  &      43.0  &      67.5  &      48.6  &      56.7  &      12.6  &      50.9  &      42.6  &      59.4  &      52.9  &      65.0  &      44.8  &      41.3  &      51.1  &      33.7  &      44.4  &      33.2  &      48.0  \\

AF-SS (ECCV'16)~\cite{qi2016augmented} &          - &          - &          - &          - &          - &          - &          - &          - &          - &          - &          - &          - &          - &          - &          - &          - &          - &          - &          - &          - &          - &      52.7  \\

SEC (ECCV'16)~\cite{kolesnikov2016seed} &      83.5  &      56.4 &       28.5 &       64.1 &       23.6 &       46.5 &       70.6 &       58.5 &       71.3 &       23.2 &         54.0 &         28.0 &       68.1 &       62.1 &         70.0 &         55.0 &       38.4 &         58.0 &       39.9 &       38.4 &       48.3 &       51.7 \\

CBTS (CVPR'17)~\cite{roy2017combining} &      85.7  &      58.8  &      30.5  &      67.6  &      24.7  &      44.7  &      74.8  &      61.8  &      73.7  &      22.9  &      57.4  &      27.5  &      71.3  &      64.8  &      72.4  &      57.3  &      37.0  &      60.4  &      42.8  &      42.2  &      50.6  &      53.7  \\

AE-PSL (CVPR'17)~\cite{wei2017object} &          - &          - &          - &          - &          - &          - &          - &          - &          - &          - &          - &          - &          - &          - &          - &          - &          - &          - &          - &          - &          - &      55.7  \\

\hline
\textit{Ours}:\\
MCOF-VGG16 &   86.8  &      73.4  &      26.6  &      60.6  &      31.8  &      56.3  &      76.0  &      68.9  &      79.4  &      18.8  &      62.0  &      36.9  &      74.5  &      66.9  &      74.9  &      58.1  &      44.6  &      68.3  &      36.2  &      64.2  &      44.0  &      \textbf{57.6}  \\

MCOF-ResNet101  & 88.2  &      80.8  &      31.4  &      70.9  &      34.9  &      65.7  &      83.5  &      75.1  &      79.0  &      22.0  &      70.3  &      31.7  &      77.7  &      72.9  &      77.1  &      56.9  &      41.8  &      74.9  &      36.6  &      71.2  &      42.6  &      \textbf{61.2}  \\

\Xhline{1.0pt}
\end{tabular} 
\vspace{0mm}
\caption{Comparison of weakly supervised semantic segmentation methods on PASCAL VOC 2012 \textit{test} set.}
\label{tab:test}
\vspace{-2mm}
\end{table*}

%We also show some failure cases in Figure~\ref{fig:failure}. For object which is too big and the key parts of it are out of the image (\textit{e.g} row 1), our method may ignore it since the key part is invisible, and the initial object seeds are missing in the training process. Fully-supervised model can partly solve this issue (Figure~\ref{fig:failure} (c)). For objects with very fine structure (\textit{e.g} bike in row 2), our method may fail due to the region segmentation and the pooling in the networks. Fully-supervised model can't address this either since the pooling reduces the resolution of features (Figure~\ref{fig:failure} (c)).

\subsection{Comparison with State-of-the-art Methods}
\begin{figure}[t] %\footnotesize
\begin{center}
\includegraphics[width=1\linewidth,trim = 0mm 0mm 0mm 0mm, clip]{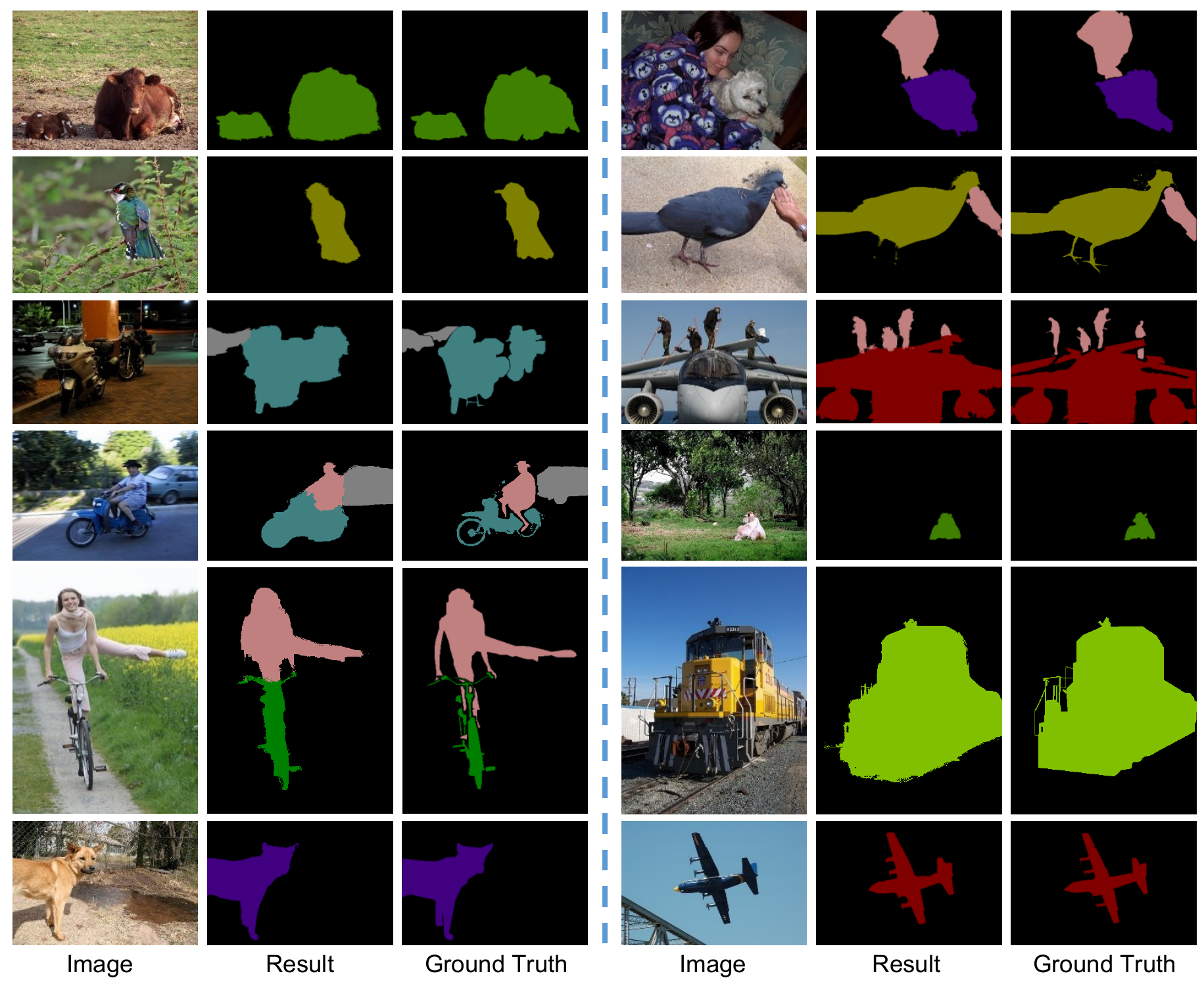}
\vspace{-5mm}
\caption{Qualitative segmentation results of the proposed framework on PASCAL VOC 2012 \textit{val} set.}
\label{fig:Qualitative}
\end{center}
\vspace{-10mm}
\end{figure}

\hyphenation{comparison}
We compare our method with previous state-of-the-art image-level weakly-supervised semantic segmentation methods: CCNN~\cite{pathak2015constrained}, EM-Adapt~\cite{papandreou2015weakly}, MIL-sppxl~\cite{pinheiro2015image}, STC~\cite{wei2016stc}, DCSM~\cite{shimoda2016distinct}, BFBP~\cite{saleh2016built}, AF-SS~\cite{qi2016augmented}, SEC~\cite{kolesnikov2016seed}, CBTS~\cite{roy2017combining} and AE-PSL~\cite{wei2017object}. 
As we mentioned above, our \textit{PixelNet} is completely compatible with fully-supervised framework and thus we can benefit from existing fully-supervised architecture. In this paper, we utilize DeepLab-LargeFOV~\cite{chen2014semantic} built on top of both VGG16 and ResNet101 as \textit{PixelNet}.
Table~\ref{tab:val} and Table~\ref{tab:test} show the comparison on mIoU on PASCAL VOC 2012 validation and test sets, respectively. We can see that our method outperforms previous methods by a large margin and achieves new state-of-the-art. When using VGG16 as basic network (MCOF-VGG16), our method outperforms the second best method, AE-PSL~\cite{wei2017object} by 1.2\% and 1.9\% on \textit{val} and \textit{test} sets, respectively. And when using the more powerful ResNet101  (MCOF-ResNet101), the improvement can reach 5.3\% and 5.5\%, respectively.
For the training samples, MIL-sppxl~\cite{pinheiro2015image} used 700K images and STC~\cite{wei2016stc} used 50K images, our method and other methods use 10K images. 
%Note that CCNN~\cite{pathak2015constrained}, EM-Adapt (Weak)~\cite{papandreou2015weakly} and SEC~\cite{kolesnikov2016seed} use 10,582 images from \textit{trainaug} set~\cite{hariharan2011semantic} in PASCAL VOC 2012 dataset. MIL+ILP+SP-sppxl~\cite{pinheiro2015image} uses 760K images from ImageNet and STC~\cite{wei2016stc} uses 41,625 simple images from Flickr-Clean dataset~\cite{wei2016stc} and 10,582 images from PASCAL VOC 2012 \textit{trainaug} set. We only use 3,770 images selected from PASCAL VOC 2012 \textit{train} set for classification task. Our method outperforms others on both validation and test sets with much fewer training samples, which demonstrates the effectiveness of the proposed iterative feature mining and network training framework. Besides, SEC~\cite{kolesnikov2016seed} mainly benefits from the additional loss: classification loss, semi-supervised loss and boundary-aware loss, while our method only used the basic SoftmaxwithLoss in Caffe framework~\cite{jia2014caffe}. The performance is benefited from the iterative framework in which more robust common features of objects are learned with iterations to predict more and more accurate segmentation results. 
We also show some qualitative segmentation results of the proposed framework in Figure~\ref{fig:Qualitative}, we can see that our weakly-supervised method can produce quite satisfactory segmentation, even in complex images. 

\subsection{Ablation Studies}
\subsubsection{Progressive Common Object Features Mining and Network Training Framework} \label{subsec:itr}
To evaluate the effectiveness of the proposed progressive common object features mining and network training framework, we evaluate the \textit{RegionNet} and \textit{PixelNet} of each iteration on training and validation set. In the ablation studies, we use VGG16 as base network for \textit{PixelNet}.
The results are shown in Table~\ref{tab:resIter}.
% We can see that the initial object seeds are very coarse (14.46\% mIoU on \textit{train} set)
%We show intermediate results in Table~\ref{tab:resIter} to evaluate the effectiveness of the proposed iterative feature mining and network training framework. 
We can see that the initial object seeds are very coarse (14.27\% mIoU on \textit{train} set), by applying the \textit{RegionNet} to learn the common features of objects, the performance achieves 29.1\%, by introducing saliency-guided refinement, it achieves 34.8\%, and after learning with the \textit{PixelNet}, it achieves 48.4\%. And in the later iterations, the performance improves gradually, which demonstrates that our method is effective.

%Note that the motivation of incorporating saliency maps is to supplement more object regions to initial object seeds. After the first iteration, the segmented masks contain much more object regions and are more accurate, while the accuracy of saliency maps are also limited, so in the later iterations, the saliency maps are not used to prevent introducing additional noise.

\begin{table}[!t]\small \setlength{\tabcolsep}{5pt}
\begin{tabular}{p{1cm}p{4cm}ccc}
\Xhline{1.0pt}
           &            &      \textit{train} &        \textit{val} \\
\Xhline{1.0pt}
           & Initial Object Seeds &      14.27 &          - \\
\hline
\multicolumn{ 1}{c|}{iter1} & \textit{RegionNet} &       29.1 &          - \\

\multicolumn{ 1}{c|}{} & \textit{Saliency-guided refinement} &       34.8 &       - \\

\multicolumn{ 1}{c|}{} & \textit{PixelNet} &       48.4 &       44.4 \\

\hline
\multicolumn{ 1}{c|}{iter2} & \textit{RegionNet} &       53.8 &          - \\

\multicolumn{ 1}{c|}{} & \textit{PixelNet} &       57.9 &       51.6 \\
\hline
\multicolumn{ 1}{c|}{iter3} & \textit{RegionNet} &       58.2 &          - \\

\multicolumn{ 1}{c|}{} & \textit{PixelNet} &       60.9 &       53.3 \\
\hline
\multicolumn{ 1}{c|}{iter4} & \textit{RegionNet} &       61.1 &          - \\

\multicolumn{ 1}{c|}{} & \textit{PixelNet} &       63.1 &       55.5 \\
\hline
\multicolumn{ 1}{c|}{iter5} & \textit{RegionNet} &       62.5     &          - \\

\multicolumn{ 1}{c|}{} & \textit{PixelNet} &       63.2 &       56.2 \\

\Xhline{1.0pt}
\end{tabular}  
\vspace{2mm}
\caption{Results of the iteration process. We evaluate the \textit{RegionNet} and \textit{PixelNet} of each iteration on training and validation sets of PASCAL 2012 dataset.}
\label{tab:resIter}
\vspace{-4mm}
\end{table}

\subsubsection{Comparison with Direct Iterative Training}
We extensively conduct experiments to verify effectiveness of the proposed progressive common object features mining and network training framework by comparing with direct iterative training method. For the direct iterative training method, we start from the segmentation results of our first iteration, and then in later iterations, use the segmentation masks of the previous iteration to train the segmentation network.

\begin{figure} %\footnotesize
\begin{center}
\includegraphics[width=1\linewidth,trim = 0mm 0mm 0mm 0mm, clip]{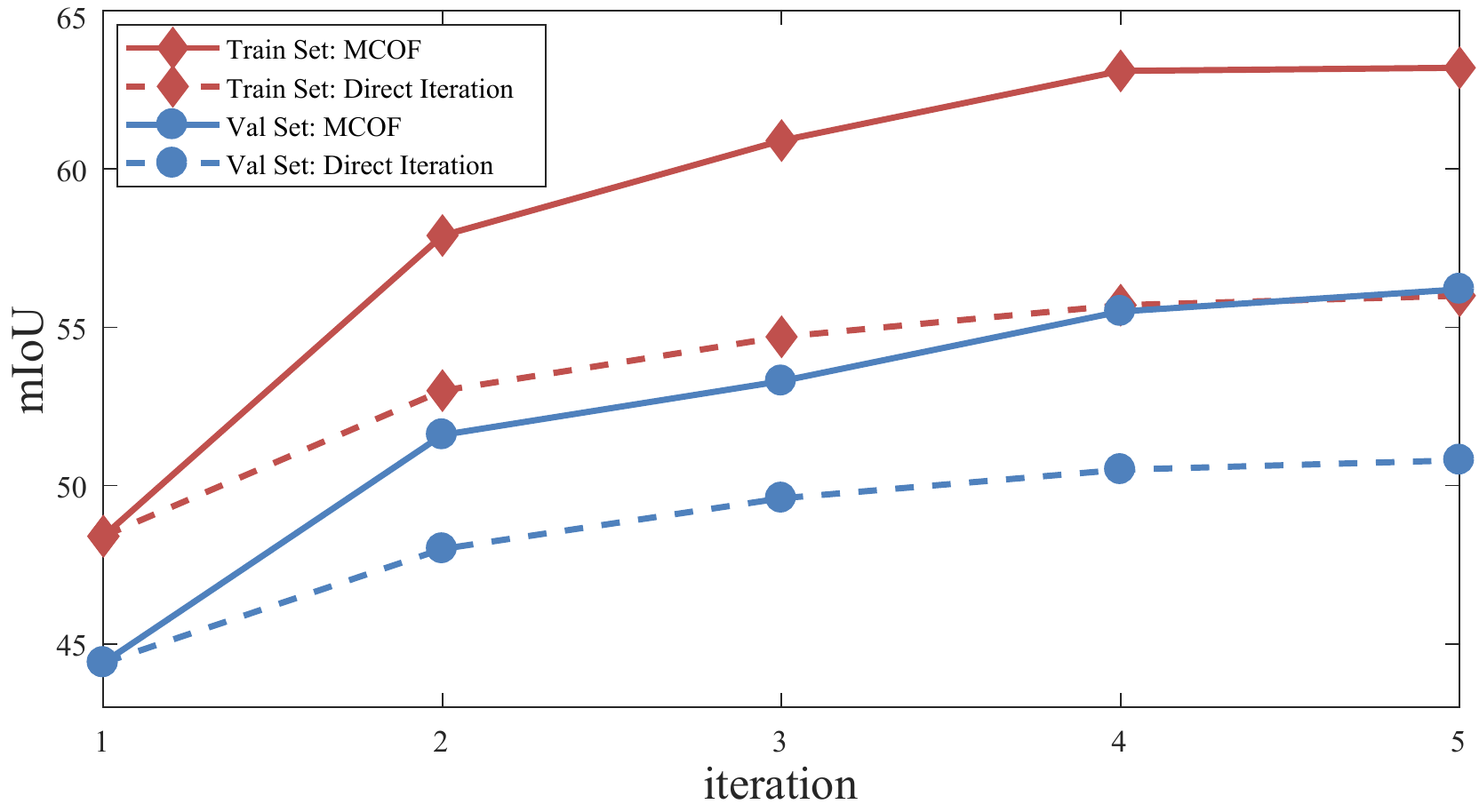} 
\vspace{-5mm}
\caption{Comparison with direct iterative training method. Our performance improves rapidly while performance of the direct iterative training method increases slowly and only reaches a low accuracy.}
\label{fig:direct_iteration}
\end{center}
\vspace{-4mm}
\end{figure}

Figure~\ref{fig:direct_iteration} shows the comparison. With the iterations, the performance of the direct iterative method increases slowly and only reaches a low accuracy, while 
in the proposed \textit{MCOF}, the performance increases rapidly and achieves much higher accuracy. This result demonstrates that our \textit{MCOF} framework is effective. The \textit{MCOF} progressively mines common object features from previous object masks and then to expand more reliable object regions to optimize the semantic segmentation network, thus the accuracy can increase rapidly to a quite satisfactory results.

%our performance improves gradually while performance of the direct iterative training method begins to decrease after 2 iterations. The reason is the direct iterative training may cause the network overfitting, thus the performance may decrease after several iterations. 
%Our method iteratively mines common object features and trains segmentation networks, they are complementary to each other, so the performance can improve gradually.

\subsubsection{Effectiveness of Saliency-Guided Refinement}
The initial object seeds only locate discriminative regions of objects, for example, heads and hands of a person, while other regions, such as the body, are often ignored.
To supplement other object regions, saliency maps are incorporated with initial object seeds. This is very important for mining the whole regions of objects. To evaluate the effectiveness, we conduct experiment on framework without saliency-guided refinement, and compare the performance of the \textit{PixelNet} of each iteration. The result is shown in Table~\ref{tab:nosaliency}. Without incorporating saliency maps, some object regions will be missing and thus the performance will be limited and can not reach satisfactory accuracy.

\begin{table}[t]\small \setlength{\tabcolsep}{6pt}

\begin{tabular}{r|ccccc}
\Xhline{1.0pt}
iterations &          1 &          2 &          3 &          4 &          5 \\
\hline
w/o saliency refinement &       41.8 &       46.2 &       47.7 &      51.5 &      52.1      \\

w/ saliency refinement &       \textbf{44.4} &       \textbf{51.6} &       \textbf{53.3} &       \textbf{55.5} &       \textbf{56.2} \\
\Xhline{1.0pt}
\end{tabular}
\vspace{-1mm}
\caption{Evaluate the effectiveness of saliency-guided refinement. We show the mIoU of the \textit{PixelNet} of each iteration on Pascal VOC 2012 val set. Without saliency-guided refinement, the performance will be limited and can not reach satisfactory accuracy.}
\label{tab:nosaliency}  
\vspace{-5mm}
\end{table}

\section{Conclusion}\label{sec:conclusion}
In this paper, we propose \textit{MCOF}, an iterative bottom-up and top-down framework which tolerates inaccurate initial localization by iteratively mining common object features from object seeds. Our method progressively expands object regions and optimizes segmentation network. In bottom-up step, starting from coarse but discriminative object seeds, we mine common object features from them to expand object regions. To supplement non-discriminative object regions, saliency-guided refinement method is proposed. Then in top-down step, these regions are used as supervision to train the segmentation network and predict segmentation masks. The predicted segmentation masks contain more complete object regions than initial, so we can further mine common object features from them. And the processes are conducted iteratively to progressively correct inaccurate initial localization and produce more accurate object regions for semantic segmentation. Our bottom-up and top-down framework bridges the gap between high-level semantic and low-level appearance in weakly-supervised semantic segmentation, and achieves new state-of-the-art performance.

\vspace{3mm}
\noindent\textbf{Acknowledgement.} This work is supported by National Key Basic Research Program of China (No. 2016YFB0100900) and National Natural Science Foundation of China (No. 61171113 and No.61773231).

\clearpage
{\small
\bibliographystyle{ieee}
\bibliography{egbib}
}

\end{document}